\documentclass[sigconf]{acmart}
\usepackage{multirow}
\usepackage{graphicx}
\renewcommand\footnotetextcopyrightpermission[1]{}

\AtBeginDocument{%
  }
\setcopyright{none} 
\acmConference{}{}{}
\acmBooktitle{}
\acmYear{}
\copyrightyear{}
\acmDOI{}
\acmISBN{}

\begin{document}

\title{GAZELOAD: A Multimodal Eye-Tracking Dataset for Mental Workload in Industrial Human–Robot Collaboration}

\author{Bsher Karbouj}
\authornotemark[1]
\email{karbouj@tu-berlin.de}
\affiliation{%
  \institution{Department of industrial Automation Technology\\
  TU Berlin}
  \city{Berlin}
  \country{Germany}
}

\author{Baha Eddin Gaaloul}
\affiliation{%
  \institution{Department of industrial Automation Technology\\
  TU Berlin}
  \city{Berlin}
  \country{Germany}
}
\author{Jörg Krüger}
\affiliation{%
  \institution{Department of industrial Automation Technology\\
  TU Berlin}
  \city{Berlin}
  \country{Germany}
}
\renewcommand{\shortauthors}{Karbouj et al.}

\begin{abstract}
    This article describes GAZELOAD, a multimodal dataset for mental workload estimation in industrial human–robot collaboration. The data were collected in a laboratory assembly testbed where 26 participants interacted with two collaborative robots (UR5 and Franka Emika Panda) while wearing Meta ARIA gen 1 smart glasses. The dataset time-synchronizes eye-tracking signals (pupil diameter, fixations, saccades, Eye Gaze, Gaze Transition Entropy, Fixation Dispersion Index) with environmental real-time/continuous measurements (Illuminance) and task/robot context (bench, task block, induced faults), under controlled manipulations of task difficulty and ambient conditions. For each participant and workload-graded task block, we provide CSV files with ocular metrics aggregated into 250 ms windows, environmental logs, and self-reported mental workload ratings on a 1–10 Likert scale, organized in participant-specific folders alongside documentation. These data can be used to develop and benchmark algorithms for mental workload estimation, feature extraction, and temporal modelling in realistic industrial HRC scenarios, and to investigate the influence of environmental factors such as lighting on eye-based workload markers.
  \end{abstract}



\maketitle

\section{Introduction} 
The increasing demand for flexibility in modern manufacturing environments has driven the integration of more intelligent and adaptive robotic systems into human–robot collaboration (HRC) \cite{zhang}. While such systems can enhance efficiency and productivity on the shop floor, their adaptive behaviors require humans to interpret system intentions, anticipate changes, and adjust their own actions accordingly, thereby increasing cognitive demands \cite{karbouj}. Against this backdrop, Industry~5.0 emphasizes human-centric and sustainable production, making the regulation of mental workload (MWL) in HRC essential, since inappropriate levels of MWL can compromise operator safety, human well-being, and overall system performance \cite{Carissoli}.
In HRC scenarios, mental workload (MWL) is commonly assessed through physiological and behavioral indicators such as eye-tracking metrics, which provide real-time and non-intrusive access to cognitive states \cite{Upasani}, or through complementary methods such as electroencephalography (EEG) or heart-rate variability (HRV) \cite{Diarra}. Recent advances in wearable devices, such as smart glasses, further support this assessment by enabling continuous, unobtrusive measurements in naturalistic industrial environments. To enable reliable MWL estimation, however, suitable datasets are required.
While numerous datasets with eye-tracking data exist, they are typically collected in general-purpose domains (e.g., driving, gaming, or office work) and are rarely available as open-access resources.
Because eye-based markers sensitive to cognitive load strongly depend on task and environmental context (e.g., illumination and noise) and on the live dynamics of interaction with the robot, such general-purpose datasets often fail to transfer accurately to industrial collaborative scenarios due to ecological-validity issues and domain shift. Therefore, the development of a domain-specific dataset for HRC scenarios is essential to enable valid estimation of mental workload under realistic, robot-interactive conditions.
Despite the growing interest in workload-aware human–robot collaboration, there is still no openly accessible dataset that captures eye-based indicators of MWL together with task and robot-interaction context in real collaborative assembly tasks.  
The primary objective of this paper is therefore to describe and share such a domain-specific dataset, providing transparent documentation of its collection protocol, structure, and modalities to enable reproducible research and method benchmarking.
We provide a multimodal dataset for estimating mental workload in human–robot collaboration. Using Meta~ARIA smart glasses, recordings were obtained during collaborative assembly tasks involving a UR5 and a Franka~Emika~Panda. The scenarios combined routine operations with pre-defined fault conditions, eliciting varying cognitive demands through error handling, corrective interventions, and coordination with the robots, while environmental context (illumination) was recorded for ecological validity.  


\section{Related Work}
Open datasets have increasingly supported research on eye-tracking as a marker of mental workload, yet their applicability to human–robot collaboration (HRC) remains limited. Pillai et al. \cite{PILLAI2020106389} released one of the first datasets, combining gaze, pupil diameter, and Likert-scale scores in a driving simulator with N-back tasks. While valuable for demonstrating workload sensitivity of ocular metrics, it remains tied to driving-specific conditions.

The COLET dataset by Ktistakis et al. \cite{KTISTAKIS2022106989} addresses the scarcity of open resources by providing raw gaze, pupil, and blink data from 47 participant performing visual search puzzles, annotated with subjective workload levels. Analyses confirm that eye features can classify workload states, but the dataset is confined to screen-based tasks and excludes embodied or collaborative contexts.

EM-COGLOAD, introduced by Miles et al. \cite{Miles}, extends this line by including 75 adults in dual-task paradigms with high/low workload labels, highlighting both load effects and age-related differences in gaze patterns. However, it too lacks interactive collaboration or robotic elements.

In summary, current datasets either offer workload labels in simulated, non-robotic domains or provide robotic interaction data without workload annotation. To date, no open dataset captures validated workload measures through eye-tracking in industrial HRC, underscoring a significant research gap that this work seeks to fill.
\section{Data Collection}
\subsection{Experimental Setup}
\begin{figure}[htb!]
    \centering
    \includegraphics[scale=1]{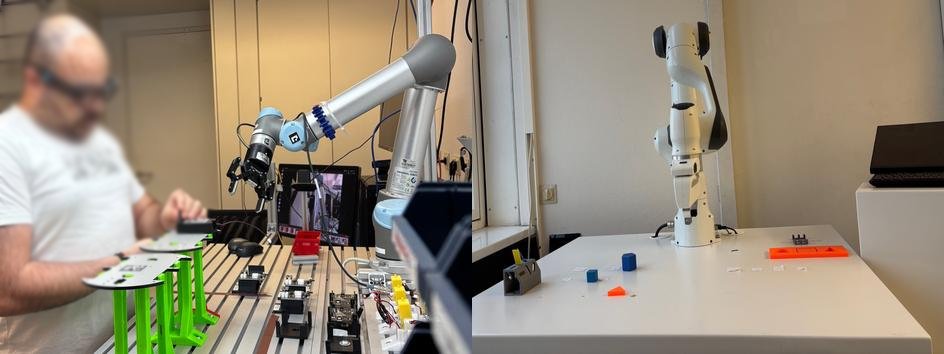}
    \caption{Experimental setup:
participant wearing ARIA smart glasses interacting with the UR5 robot (left);
Franka~Emika~Panda station with assembly parts (right)}
    \label{fig:setup}
\end{figure}

The experimental study was conducted in a laboratory testbed designed to emulate a manufacturing assembly environment. Two collaborative robots were employed: a UR5 (Universal Robots) and a Franka Emika Panda, each mounted on dedicated workstations for performing standardized assembly operations. Human participant were equipped with Meta ARIA smart glasses, which provided multimodal recordings including eye-tracking data, egocentric video, inertial measurement unit (IMU) data, and audio. In addition, environmental illumination (lux) was continuously recorded using a high-precision light sensor mounted on the smart glasses to account for contextual influences on cognitive load. All robot actions and task progressions were logged in synchronization with participant responses, enabling multimodal alignment of physiological signals, environmental conditions, and task events. Figure~\ref{fig:setup} illustrates the physical setup with the participant equipped with ARIA glasses.
\subsection{participant}  
Twenty-six participants were recruited from the university community (16 male, 10 female; mean age 23.3 years, SD 2.6, range 20-34). All reported normal or corrected-to-normal vision; color blindness and ocular conditions were exclusion criteria. 
Each experimental session lasted approximately 60 minutes. All participants were adults and received an explanation of the study goals and procedures prior to the experiment. Written informed consent was obtained for participation and for the collection and use of eye-tracking and video data (signed privacy policy). All data were anonymized before storage and analysis, and no personally identifying raw video is released.
\begin{figure*}[htb!]
    \centering
    \includegraphics[scale=0.8]{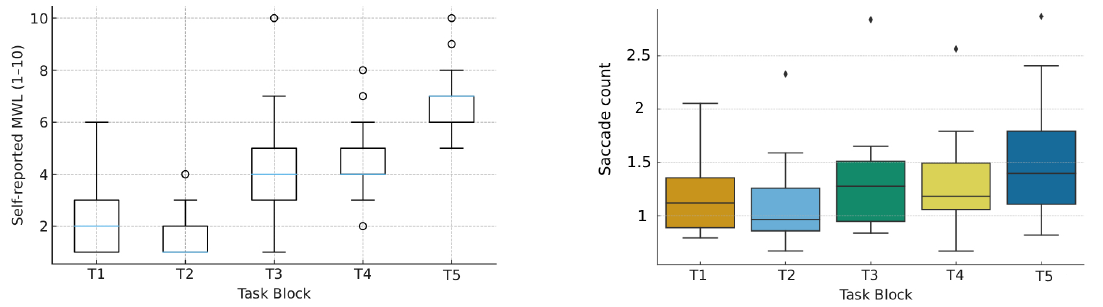}
    \caption{Example summaries from the dataset: (left) self-reported MWL (1–10) per task block T1–T5 aggregated across participant; (right) saccade count per 250\,ms window per task block.}
    \label{fig:summary_boxplots}
\end{figure*}
\subsection{Experimental Tasks}
During the experiment, participants alternated between two roles: (i) passive monitoring of robot behavior and (ii) active collaboration to correct or guide the robots in the presence of faults. These roles were chosen to span a range of cognitive demands from situation monitoring to problem-solving and coordination. The protocol comprised three workload-graded task blocks:
\begin{itemize}
    \item \textbf{Task 1 (low load):} Participants interacted sequentially with the UR5 and Franka benches. At the UR5 station, the robot executed an error-free assembly sequence on real components, while participants observed the process and monitored for deviations from expected behavior. At the Franka station, participants performed a precision-sensitive assembly step with real components, where the accuracy of part placement determined whether the robot could proceed correctly. No artificial preset/preprogrammed robot errors were injected at this bench; any disturbances arose from imprecise human actions that occasionally required corrective guidance. After completing each part at both benches, participants provided a self-rating of perceived mental workload on a 1–10 Likert scale. In the Dataset, Task 1 (low load) refers to sessions 1 and 2 (UR5 and Franka respectively).
    \item \textbf{Task 2 (medium load):} Again, two parts were performed at the UR5 and Franka benches, but participants now played an interactive role throughout. At the UR5 station, the robot executed an assembly sequence that included pre-simulated faults (e.g., incorrect placement or missing actions causing the control to beep loudly). Participants had to detect, diagnose, and correct these faults through direct input. At the Franka bench, robot behavior was likewise perturbed by predefined error conditions, requiring participants to intervene and provide corrective guidance. After each part, participants rated their perceived MWL on the same 1–10 scale. This block was designed to increase cognitive demands relative to Task 1 by introducing systematic fault detection and resolution in real assembly contexts. In the Dataset, Task 2 (medium load) refers to sessions 3 and 4 (UR5 and Franka respectively).
    \item \textbf{Task 3 (high load):} The third block integrated the two benches into a single continuous sequence intended to elicit the highest workload level. Participants alternated between the spatially separated UR5 and Franka stations, introducing context-switching and navigation demands. At both benches, pre-simulated errors were triggered in overlapping time intervals, forcing participants to divide attention and coordinate interventions across two different robot platforms and task types. At the UR5 station, participants monitored and corrected faulty robot actions at the Franka station, they executed precision-dependent assembly steps subject to induced disturbances. After completing each part, participants again reported their perceived MWL on the 1–10 scale. In the Dataset, Task 3 (high load) refers to session 5.
\end{itemize}
To prevent carry-over and saturation effects, a standardized break of approximately three minutes was scheduled between task blocks as well as a standardized break of approximately two minutes was scheduled between both parts of the same task block.
\section{Dataset}
The dataset is organized into four main folders: \textit{Metadata}, \textit{Lux Data}, \textit{Eventlogger Data}, and \textit{Eye Metrics}. The \textit{Metadata} folder contains participant demographics and subjective ratings in Excel format (\texttt{Tasks\_Rating.xlsx}). The \textit{Lux Data}, \textit{Eventlogger Data}, and \textit{Eye Metrics} folders each contain 130 CSV files, corresponding to the complete experimental design of 26 participants $\times$ 5 experimental sessions per participant.

All files follow a consistent naming convention: \texttt{XX\_Y\_DataType}, where \texttt{XX} represents the participant identifier (01--26) and \texttt{Y} represents the session number (1--5). Specifically, files are named \texttt{XX\_Y\_lux.csv} for real-time ambient light measurements, \texttt{XX\_Y\_eventlogger.csv} for behavioral event logs, and \texttt{XX\_Y\_Metrics\_withLux.csv} for eye-tracking metrics with integrated lux data. This uniform nomenclature enables straightforward cross-referencing across data modalities and facilitates automated batch processing.

All three data streams use common timestamp fields for precise temporal alignment. To enable precise multimodal synchronization, the lux and eventlogger files store timestamps in ISO~8601 UTC format, whereas the eye-metrics files provide \texttt{timestamps\_ms} (milliseconds since session start); all streams are aligned using UTC as the common temporal reference. 

The eye-metrics files employ milliseconds since session start, while the lux files and eventlogger files use ISO~8601 datetime format. This temporal synchronization allows researchers to simultaneously analyze eye-tracking dynamics, behavioral events, and real-time ambient light conditions within the same experimental session. Supporting documentation at the repository root level includes processing scripts (\texttt{metrics\_extraction.py}, \texttt{pupil.py}). Table~\ref{tab:dataset_overview} summarizes all dataset components, key variables, measurement units, and sampling rates.

\begin{table*}[htb!]
\centering
\caption{Comprehensive overview of the released dataset components.}
\label{tab:dataset_overview}
\resizebox{\textwidth}{!}{
\begin{tabular}{p{3.0cm} p{3.2cm} p{5.2cm} p{6.0cm} p{2.4cm} p{3.0cm}}
\toprule
Category & File name & Key variables & Description & Unit / scale & Sampling \\
\midrule

\textbf{Meta Data} 
& \texttt{Tasks\_Rating.xlsx} 
& \texttt{participant\_id}, \texttt{sex (m/f/d)}, \texttt{age}, \texttt{tasks} 
& Participant demographics and task/session annotations 
& -- 
& Manual input \\
\cline{2-6}

& \texttt{metrics\_extraction.py} 
& -- 
& Python script for ocular-metric extraction and aggregation 
& -- 
& -- \\
\cline{2-6}

& \texttt{pupil.py} 
& -- 
& Python script for pupil processing and feature computation 
& -- 
& -- \\

\midrule
\textbf{Lux Data} 
& \texttt{XX\_Y\_lux.csv} 
& \texttt{timestamp}, \texttt{lux\_value} 
& Continuous real-time illumination measurements recorded by a high-precision VEML7700 light sensor mounted on the ARIA smart glasses 
& lux 
& 1\,Hz raw $\rightarrow$ linearly interpolated to 4\,Hz in Eye-Metrics files for temporal alignment \\

\midrule
\textbf{Eventlogger Data} 
& \texttt{XX\_Y\_eventlogger.csv} 
& \texttt{Time\_UTC}, \texttt{Action}, \texttt{ID} 
& Manual logging of experimental events, including session start, end, and action timestamps 
& -- 
& Manual input \\

\midrule
\textbf{Eye-Metrics} 
& \texttt{XX\_Y\_Metrics\_withLux.csv} 
& \texttt{selfreport}, \texttt{task}, \texttt{timestamps\_ms}, \texttt{fixation\_count}, \texttt{saccade\_count}, 
\texttt{saccade\_amplitude\_deg}, \texttt{saccade\_velocity\_deg\_s}, 
\texttt{EyeGaze\_x}, \texttt{EyeGaze\_y}, \texttt{GTE}, \texttt{FDI}, \texttt{SaccRate}, \texttt{blink\_flag}, \texttt{lux\_interpolated} 
& Continuous binocular ocular measurements recorded by Meta ARIA smart glasses and derived metrics for fixation and saccade dynamics 
& mm, [deg], [deg/s], counts 
& 60\,Hz raw $\rightarrow$ aggregated into non-overlapping 250\,ms windows (4.0\,Hz) \\

\bottomrule
\end{tabular}
}
\vspace{2mm}
\small\textit{All timestamps serve as a common temporal reference across files to enable precise multimodal synchronization.}
\end{table*}

\begin{figure}[htb!]
    \centering
    \includegraphics[scale=0.55]{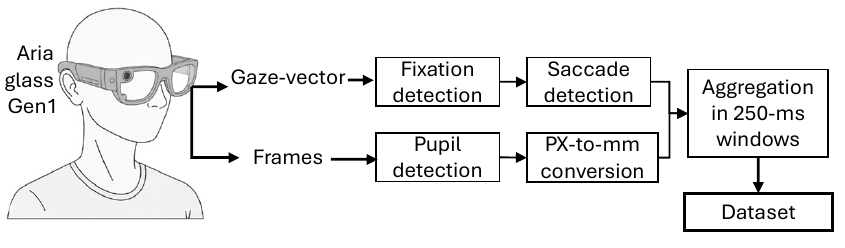}
    \caption{Preprocessing pipeline for deriving eye-tracking metrics from raw ARIA smart-glasses recordings.}
    \label{fig:pipeline}
\end{figure}

Figure~\ref{fig:pipeline} illustrates the preprocessing workflow for deriving eye-tracking metrics from the raw ARIA recordings. The ARIA Gen-1 smart-glasses provide two primary streams: the video frames from the scene cameras and the gaze-vector estimated in the headset coordinate system. From these raw signals, pupil detection was performed on the video frames followed by pixel-to-millimetre conversion using the ARIA calibration parameters. Gaze-vector samples were processed with a velocity-threshold identification (I-VT) method to detect fixations and saccades. All derived metrics were then temporally aligned and aggregated into 250-ms non-overlapping windows, resulting in a synchronized time-series suitable for workload-estimation studies.
Angular-velocity threshold for saccades is set to 30~[deg/s], minimum fixation duration to 60~ms, and a 75~ms merge gap is used to join adjacent fixations. Pixel-to-millimetre conversion uses ARIA’s per-session calibration parameters.

\section{Data availability \& Licensing}
All extracted dataset files are publicly available via \href{https://data.mendeley.com/datasets/9smd7nbtwc/1}{\textit{Mendeley Data}} and released under the Creative Commons Attribution license (CC BY~4.0). 
The repository is identified by the DOI \href{https://doi.org/10.17632/9smd7nbtwc.1}{\texttt{10.17632/9smd7nbtwc.1}}. 
The release includes the full set of CSV files (Lux, Eventlogger, and Eye-Metrics streams), supporting metadata, and processing scripts as documented in Table~\ref{tab:dataset_overview}. 
No raw personally identifying images or egocentric video are provided; only anonymized and derived signals are shared in extracted CSV format. 
Issues and requests are handled via the public tracker, and versioning and changes are logged in \texttt{CHANGELOG.md}.

\section{Ethics \& Anonymization}
All participants were adults and provided written informed consent prior to participation, including consent for the collection and anonymized sharing of their eye-tracking and behavioral data. The study involved only non-invasive observation of human–robot interaction in a laboratory setting, without any medical procedures, cognitive stress induction protocols beyond normal task difficulty, or collection of personally identifying biometric attributes. According to the institutional guidelines of the Technical University of Berlin, Germany, such non-interventional behavioral studies are formally exempt from ethics committee review.

To protect participant privacy, no personally identifying images or raw egocentric video are released. Only anonymized ocular features, derived signals, environmental measurements, task/robot logs, and self-reported workload ratings are included in the public dataset, provided as extracted data files (CSV format).

\section{Limitations}
The dataset was collected in a single laboratory using one collaborative assembly testbed with two specific cobots and a fixed workstation layout, so the recordings primarily reflect this environment and task family. Participants are healthy adults within a relatively narrow age range and are recruited from a university context, which limits demographic diversity. Only eye-based and event-logging measures are included; no additional physiological signals (e.g., EEG, ECG) or detailed motion capture are provided.







\bibliographystyle{ACM-Reference-Format}
\bibliography{sample-base}

@String{Computer = "{IEEE} Computer" }

@article{zhang,
title = {Towards new-generation human-centric smart manufacturing in Industry 5.0: A systematic review},
journal = {Advanced Engineering Informatics},
volume = {57},
pages = {102121},
year = {2023},
issn = {1474-0346},
author = {Chao Zhang and Zenghui Wang and Guanghui Zhou and Fengtian Chang and Dongxu Ma and Yanzhen Jing and Wei Cheng and Kai Ding and Dan Zhao},
}

@article{karbouj,
title = {Adaptive Behavior of Collaborative Robots: Review and Investigation of Human Predictive Ability.},
journal = {Procedia CIRP},
volume = {130},
pages = {952-958},
year = {2024},
note = {57th CIRP Conference on Manufacturing Systems 2024 (CMS 2024)},
issn = {2212-8271},
author = {Bsher Karbouj and Kotayba Al Rashwany and Obada Alshamaa and Jörg Krüger},
}

@article{Carissoli,
author = {Carissoli, Claudia and Negri, Luca and Bassi, Marta and Storm, Fabio and Delle Fave, Antonella},
year = {2023},
month = {09},
pages = {1-20},
title = {Mental Workload and Human-Robot Interaction in Collaborative Tasks: A Scoping Review},
volume = {40},
journal = {International Journal of Human-Computer Interaction},
}

@article{Upasani,
author = {Upasani, Satyajit and Srinivasan, Divya and Zhu, Qi and Du, Jing and Leonessa, Alexander},
year = {2023},
month = {10},
pages = {},
title = {Eye-Tracking in Physical Human–Robot Interaction: Mental Workload and Performance Prediction},
volume = {66},
journal = {Human Factors: The Journal of the Human Factors and Ergonomics Society},
}

@ARTICLE{Diarra,
  
AUTHOR={Diarra, Moussa  and Theurel, Jean  and Paty, Benjamin },
         
TITLE={Systematic review of neurophysiological assessment techniques and metrics for mental workload evaluation in real-world settings},
        
JOURNAL={Frontiers in Neuroergonomics},
        
VOLUME={Volume 6 - 2025},

YEAR={2025},

}

@article{PILLAI2020106389,
title = {Response time and eye tracking datasets for activities demanding varying cognitive load},
journal = {Data in Brief},
volume = {33},
pages = {106389},
year = {2020},
issn = {2352-3409},
author = {Prarthana Pillai and Prathamesh Ayare and Balakumar Balasingam and Kevin Milne and Francesco Biondi},
}

@article{KTISTAKIS2022106989,
title = {COLET: A dataset for COgnitive workLoad estimation based on eye-tracking},
journal = {Computer Methods and Programs in Biomedicine},
volume = {224},
pages = {106989},
year = {2022},
issn = {0169-2607},
author = {Emmanouil Ktistakis and Vasileios Skaramagkas and Dimitris Manousos and Nikolaos S. Tachos and Evanthia Tripoliti and Dimitrios I. Fotiadis and Manolis Tsiknakis},
}

@article{Miles,
author = {Miles, Gabriella and Smith, Melvyn and Zook, Nancy and Zhang, Wenhao},
year = {2024},
month = {03},
pages = {},
title = {EM-COGLOAD: An investigation into age and cognitive load detection using eye tracking and deep learning},
volume = {24},
journal = {Computational and Structural Biotechnology Journal},
}
\end{document}